\documentclass[acmsmall]{acmart}
\usepackage{hyperref}
\usepackage{multirow}
\usepackage{colortbl}
\usepackage{xcolor}
\AtBeginDocument{%
  }

\usepackage{booktabs}
\usepackage[linesnumbered,ruled,vlined]{algorithm2e}
\usepackage{enumitem}
\usepackage{rotating}

\begin{document}

\title{PEAS: A Strategy for Crafting \\Transferable Adversarial Examples}


\author{Bar Avraham}
\affiliation{%
  \institution{Ben-Gurion University}
  \country{Israel}
 }
 
\author{Yisroel Mirsky}
\affiliation{%
  \institution{Ben-Gurion University}
  \country{Israel}
 }
  \authornote{Corresponding Author}

\renewcommand{\shortauthors}{Bar Avraham and Yisroel Mirsky}

\begin{abstract}
Black box attacks, where adversaries have limited knowledge of the target model, pose a significant threat to machine learning systems. Adversarial examples generated with a substitute model often suffer from limited transferability to the target model. While recent work explores ranking perturbations for improved success rates, these methods see only modest gains. We propose a novel strategy called PEAS that can boost the transferability of existing black box attacks. PEAS leverages the insight that samples which are perceptually equivalent exhibit significant variability in their adversarial transferability. Our approach first generates a set of images from an initial sample via subtle augmentations. We then evaluate the transferability of adversarial perturbations on these images using a set of substitute models. Finally, the most transferable adversarial example is selected and used for the attack. Our experiments show that PEAS can double the performance of existing attacks, achieving a 2.5x improvement in attack success rates on average over current ranking methods. We thoroughly evaluate PEAS on ImageNet and CIFAR-10, analyze hyperparameter impacts, and provide an ablation study to isolate each component's importance.

\end{abstract}

\maketitle
\section{Introduction}

Adversarial examples are subtly altered inputs that mislead machine learning models. These samples pose a significant threat to the security of AI systems. Of particular concern are black box attacks, where the adversary lacks detailed knowledge of the target model's architecture or parameters. This scenario reflects the reality of most commercial AI systems deployed in the cloud or embedded in products where the adversary can only interact with the model through queries.

A common strategy for black box attacks relies on the use of substitute models. The adversary trains a substitute model ($f'$) and generates adversarial examples tailored to it, hoping for transferability to the target model ($f$) due to gradient alignment \cite{demontis2019adversarial}. However, adversarial transferability remains a significant challenge. Inherent differences in model architectures, training data, and optimization techniques can lead to gradients that point in vastly different directions in the input space. This mismatch between the substitute model and the target model often results in adversarial examples that are highly effective against the substitute model but fail to fool the target model.

Our key observation is that there often exist numerous samples that are perceptually equivalent to the original input ($x$), yet exhibit significant variability in their alignment with other models' decision boundaries. If the adversary can discover a perceptually equivalent sample that has good alignment with unknown models, they can substantially increase their chances of a successful attack. The challenge lies in efficiently exploring the space of perceptually equivalent samples and selecting the one most likely to transfer to the target model $f$.

\begin{figure*}
    \centering
    \includegraphics[width=\textwidth]{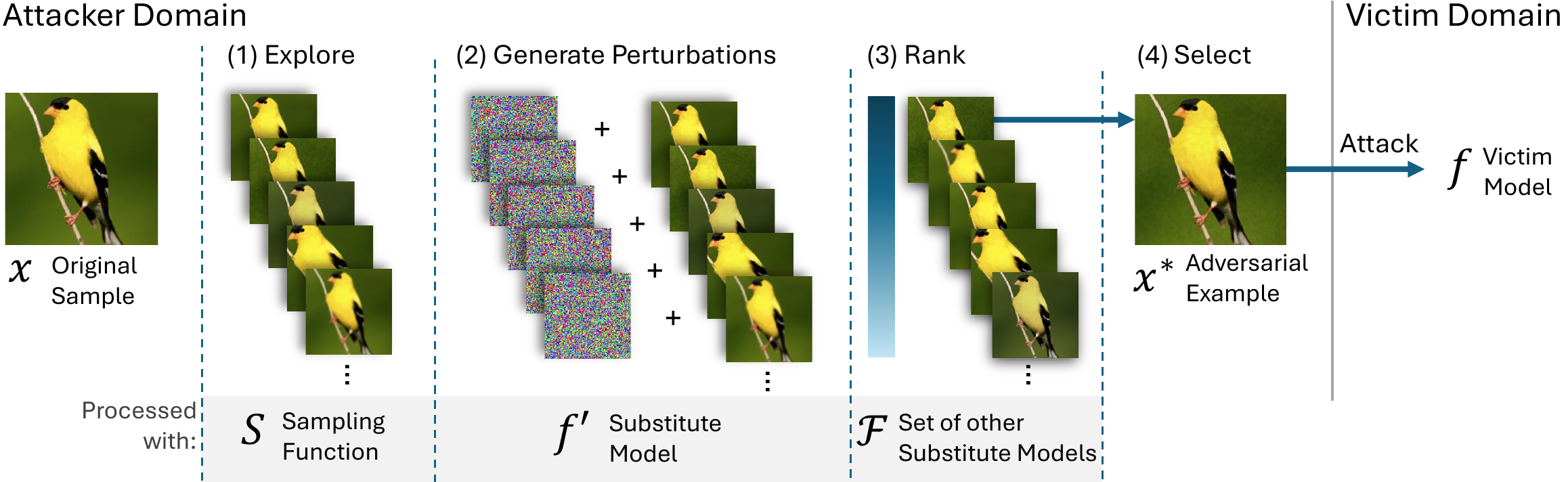}
    \caption{The attack process of PEAS: (1) explore the space around input $x$ by generating perceptually equivalent images with a sampling function (e.g., subtle image augmentations), (2) attack each sample using \textit{any} adversarial example algorithm (e.g., a white box attack on substitute model $f'$), (3) measure the expected transferability of each sample using a set of substitute models $\mathcal{F}$, (4) select the sample that has the highest expected transferability score ($x^*$) and use it for the attack on the victim's model $f$.}
    \label{fig:mainfig}
    \vspace{-1em}
\end{figure*}

We introduce the Perception Exploration Attack Strategy (PEAS), a novel method for boosting the transferability of adversarial examples, which can be applied to existing black box attacks. PEAS begins by generating a set of perceptually equivalent variations of the input $x$ using \textit{subtle} image augmentations (e.g., randomly shifting the image by a few pixels). PEAS then attacks each of these variations with a user provided attack algorithm (e.g., PGD on a substitute model $f'$ or some other black box attack algorithm). This results in a set of adversarial examples for $x$. Finally, the transferability of each of these adversarial examples is estimated using a set of substitute models ($F$), and the most transferable sample is selected for the attack. This process is illustrated in Fig. \ref{fig:mainfig}.

While adversarial perturbations are conventionally constrained by p-norms to preserve the stealth of the attack, we argue that common image transformations such as pixel shifts and slight rotations maintain stealth if performed in moderation. However, in contrast to noise-based strategies such as random start, we have discovered that image transformations result in starting points that are more likely to align with the gradients of unknown target models. We discuss these insights and implications in our work.

In this paper, we perform comprehensive evaluations and demonstrate that PEAS can achieve state-of-the-art performance in black box attack settings. We surpass the success rates of existing ranking based methods and black box attacks by a significant margin across various datasets and network architectures. Moreover, through an ablation study, we verify the contribution of each component and show that the success of the attack is directly attributed to the strategy and not due to misclassification errors from the augmentations.

In summary, this paper has the following contributions:
\begin{itemize}[itemsep=0ex]
    \item We uncover a crucial finding that there is a specific set of starting points which, if attacked, can significantly increase the adversarial example's transferability. This lays the foundation for our novel attack strategy.
    \item We introduce the concept of perceptual equivalence and discuss how perceptually equivalent images maintain the adversarial objective of stealth. 
    \item We propose two strategies for finding perceptually equivalent images using \textit{subtle} image augmentation. We explore and discuss why these images are significantly better starting points for discovering transferable adversarial examples. 
    \item We propose a framework (PEAS) that leverages these insights to boost the transferability of existing black box attacks. To the best of our knowledge, we are the first to show how transferability ranking can be used to craft \textit{effective} adversarial examples.   
    \item Using PEAS, we show that it's possible to create a black box adversarial example using subtle perturbations alone, without adversarial perturbations (noise).
\end{itemize}

\section{Related Works}
Our study introduces a novel method to boost the transferability of adversarial examples made using substitute models. We start by reviewing the concept of transferability and then examine how previous research has aimed to improve it by (1) enhancing perturbation robustness and (2) choosing the best perturbation for each sample, known as ranking.

\textbf{Transferability.} The term transferability refers to the phenomenon where adversarial examples generated using a substitute model can effectively deceive another model. This principle was first highlighted by Szegedy et al. \cite{szegedy2013intriguing} and further explored by Goodfellow et al. \cite{goodfellow2014explaining} that showed that adversarial training can alleviate transferability slightly, and by Papernot et al.  \cite{papernot2016transferability} who demonstrated the ability of adversarial perturbations to generalize across different models. The reason for this transferability is often attributed to the similarity in gradient directions or decision boundaries between the models, a phenomenon known as gradient alignment. This concept suggests that despite variations in architecture or training data, different models may still exhibit vulnerabilities to the same adversarial examples. Demontis et al. \cite{demontis2019adversarial} further reveal this concept by examining the role of gradient alignment in transferability, providing a more technical foundation for understanding why and how adversarial examples can deceive multiple models. 

With transferability, an attacker can take a sample $x$, craft an adversarial example $x'$ using an arbitrary model $f'$, and expect some level of success when using it on the victim's model $f$. However, the attack success rates in this naive transferability setting are usually quite low \cite{ozbulak2021selection}. 

\textbf{Improving Transferability.} To improve attack success rates, researchers have looked for ways to increase the likelihood of transferability. The general approach is to increase diversity in the process of creating $x'$ to simulate the loss surface and decision boundaries of unknown models \cite{bhambri2019survey}. Works such as \cite{liu2016delving,ding2021low,ma2021simulating,cai2022blackbox,lord2022attacking,feng2022boosting,qin2023training} increase model diversity by using multiple substitute models with different architectures. The idea is that if $x'$ works on a set of different models (i.e., crosses their decision boundaries), then it is likely to work on an unknown model. Other works, such as \cite{dong2018boosting} modify the optimization algorithm to mitigate the issue of overfitting to $f'$.

Another approach has been to increase input diversity to $f'$. For example, Xie et al. showed that it is possible to make a robust adversarial perturbation by applying random transformations (i.e., augmentations such as random resizing and padding) to the sample at each iteration during its generation on $f'$ \cite{xie2019improving}. This process is similar to expectation over transformation (EOT) \cite{athalye2018synthesizing} and has been used in various different ways to make transferable perturbations \cite{lin2019nesterov,zou2020improving,zhu2022attention}. Dong et al. improved the process further by applying an augmentation kernel to the perturbation itself, making the entire process more efficient and effective \cite{dong2019evading}. 

All of these works have been trying to solve the problem of making a \textit{robust perturbation} for $x$ using $f'$, whether it be by using multiple models or by performing transformations on $x$ during the optimization process. Our work aims to solve a different problem: of all the perceptually identical images to $x$, which one gives us the most advantageous starting point for creating an adversarial example with higher likelihood of transferability? 

\textbf{Measuring Transferability.} In a work done by Ozbulak et al. \cite{ozbulak2021selection}, it was discovered that certain sample subsets exhibit superior transfer capabilities. Later in \cite{levy2022transferability}, this insight was used to rank the expected transferability of a set of adversarial examples. The approach is to rank each sample according to its ability to induce uncertainty in the predictions of a set of substitute models $\mathcal{F}$. The authors found that their approach works well when ranking \textit{different images} but not so well when ranking different versions of \textit{the same image}-a feature necessary for creating adversarial examples.

We identify the root cause of this limitation: random noise from an $\epsilon$-ball often fails to perturb the robust features crucial for transferability. Therefore, we are the first to propose how transferability ranking can be used to effectively \textbf{craft adversarial examples} by overcoming this limitation.  Our key insight is that subtle augmentations to robust features are significantly more effective in exploring samples with high transferability. This novel approach yields a 2.5x average improvement in performance over \cite{levy2022transferability}. Furthermore, we present the first framework that can be applied to existing black box attacks, significantly improving their performance.

\section{Perceptual Exploration Attack Strategy (PEAS)}\label{sec:peas} 
In this section we present our novel attack strategy. First, we introduce the concept of perceptual equivalence and then discuss how it can be used in conjunction with ranking to boost the transferability of adversarial examples in black box settings.

\subsection{Perceptual Equivalence}
An adversary's goal is to generate an adversarial example $x'$ that fool a target classifier $f$ while remaining indistinguishable from the original input $x$. This stealthiness is often achieved by limiting adversarial changes to lie within an $\epsilon$-ball around $x$, as measured by a p-norm distance metric ($||x - x'||_p < \epsilon$).

However, the p-norm metric doesn't perfectly align with human perception. We can make changes to an image that drastically increase its p-norm while remaining visually imperceptible to a casual human observer. For example, by shifting an image by two pixels. Therefore, we define two images $x_i$ and $x_j$ as \textit{perceptually equivalent} if a casual human observer would deem them the same, with no suspicions about $x_j$. 

\begin{figure}[t]
    \centering
    \includegraphics[width=0.7\columnwidth]{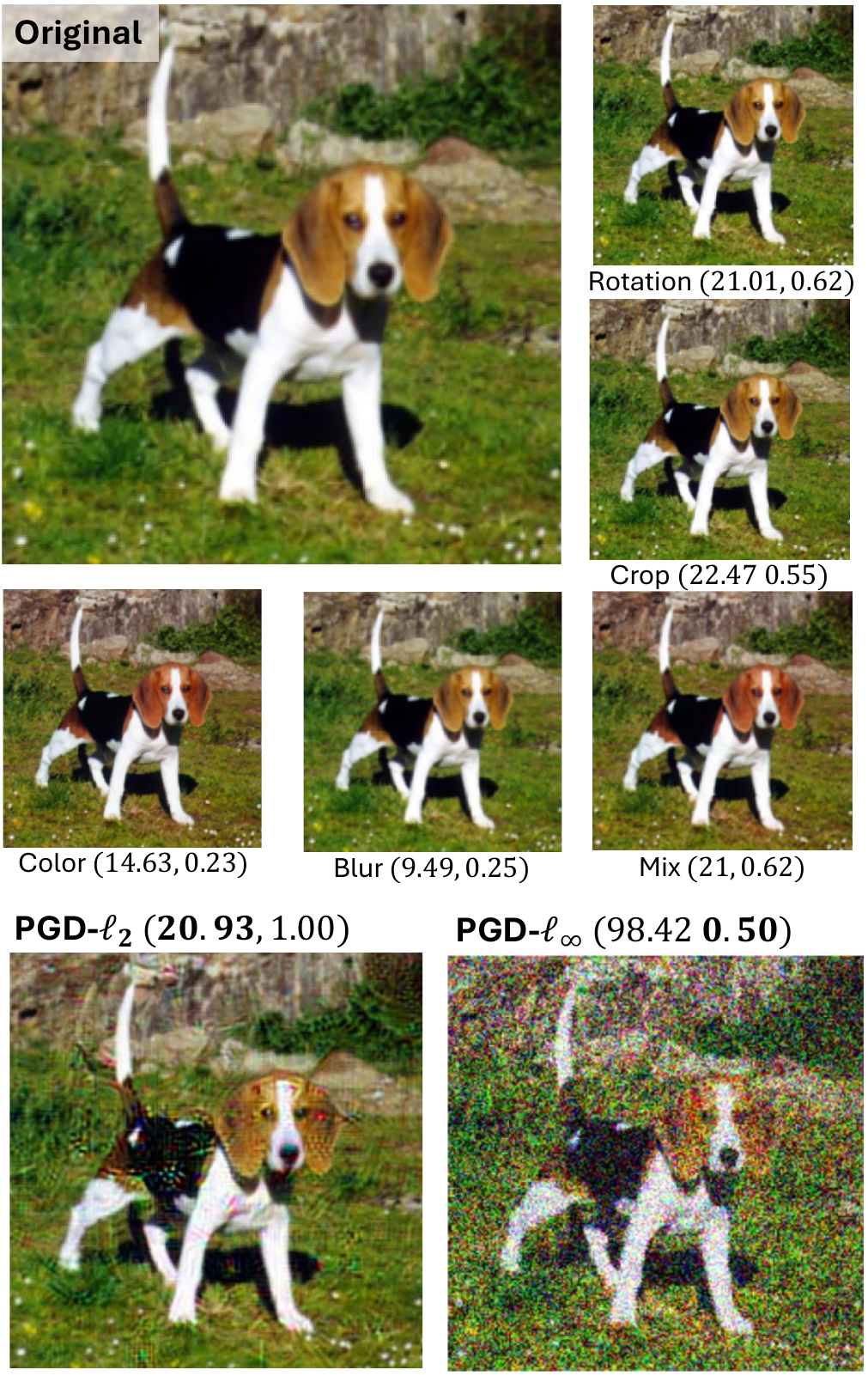}
    \caption{This example demonstrates how subtle augmentations can result in large $(\ell_2, \ell_\infty)$ distances from the original image yet remain perceptually equivalent. Therefore, we argue that these subtle transformations can be used in an adversarial example attack.}
    \label{fig:augnorms}
\end{figure}

We argue that adversaries can exploit perceptual stealth rather than relying solely on p-norm constraints.  Fig. \ref{fig:augnorms} illustrates this concept by subtly augmenting an image. While these versions seem identical to humans, their $\ell_2$ and $\ell_\infty$ norms are much higher than typical black box attack $\epsilon$-budgets. This is in contrast to adversarial perturbations of the same p-norm magnitude. For reference, a `large' p-norm distance for adversarial examples is $(1, 0.05)$ for ImageNet.

An interesting observation is that an image with a subtle augmentation has its robust features (i.e., the main features used in classification) perturbed, whereas an image with additive noise has its non-robust features (noise patterns) perturbed. In either case, an alteration to either type of feature will affect the sample's location with respect to the model's loss surface. We will now discuss the implication of this phenomenon. 

\subsection{Starting Points \& Transferability}
A common strategy for improving adversarial examples is to try running the attack multiple times from different locations near $x$ and by selecting the best result \cite{serban2020adversarial}. This strategy, known as `random starts,' is effective because different start points can lead to different optima on the loss surface of $f'$. In the context of transferability, we seek a starting point which has good gradient alignment with an unknown model $f$. 

Let $S(x)$ denote a sampling function that produces a sample near $x$. As shown by Levy et al. \cite{levy2022transferability}, among the samples produced by $S(x)$, there exists a sample which, if attacked using substitute $f'$, will exhibit superior \textit{transferability} to an unknown model $f$. We hypothesize that these starting points generalize well to other models because they are either (1) near a shared boundary or (2) have a gradient that aligns well with other models.

We build upon this insight: By employing a sampling strategy that generates samples that are perceptually equivalent to \(x\), we can enhance the probability of creating a sample with improved transferability. This is because decision boundaries are more strongly influenced by robust features than non-robust features (noise) \cite{ilyas2019adversarial} and therefore hold greater potential for placing the sample in a superior starting position compared to a randomly selected point within an  $\epsilon$-ball around $x$ (as done in \cite{levy2022transferability}). We empirically validate this claim in our evaluations and show that the added benefit is \textit{not} because the augmentations cause natural misclassifications (see section \ref{sec:eval}). 

The challenge lies in efficiently exploring samples that are perceptually equivalent to $x$ while identifying those which, when adversarially perturbed using $f'$, will exhibit the highest likelihood of transferring to $f$.

\subsection{The Attack Strategy}

The proposed perceptual exploration attack strategy is designed to systematically explore the space of perceptually equivalent variations of an input sample $x$ and select the variation that is most likely to transfer to an unknown model. The core steps are shown in Fig. \ref{fig:mainfig} and presented in Algorithm \ref{alg:peas}. The following is a detailed explanation of the process:

\begin{enumerate}[itemsep=0em]
    \item \textbf{Perceptual Exploration:} We begin by generating a set of $n$ perceptually equivalent samples to $x$. This is achieved by applying a sampling function $S$, which generates a subtly augmented version of $x$, $n$ times. Let $X = \{x_1, x_2, ..., x_n\}$ be the resulting set of augmented samples.
    \item \textbf{Adversarial Perturbation:} We attack each sample $x_i \in X$ using a substitute model $f'$. The attack is performed using \textit{any} adversarial example attack algorithm (black box in $f$ or white box on $f'$). This results in a set of adversarial examples $X' = \{x'_1, x'_2, ..., x'_n \}$.
    \item \textbf{Transferability Estimation:} Since we don't have access to the target model $f$, we estimate the transferability of each adversarial example in $X'$  using the expected transferability metric (ET) \cite{levy2022transferability}. Let $\mathcal{F}$ represent a set of substitute models (not including the model $f'$ used for attack generation).  The ET of $x'_i$ is computed as:
   \begin{equation}\label{eq:het}
       \text{ET}_\mathcal{F}(x) = \frac{1}{|\mathcal{F}|} \sum_{f \in \mathcal{F}} [1 - \sigma_y(f(x))]
   \end{equation}

   where $\sigma_y$ is the softmax output corresponding to the original class $y$ of sample $x$ (assuming an untargeted attack).  Intuitively, ET measures how often the adversarial example succeeds in fooling a diverse set of substitute models.

   \item \textbf{Sample Selection:}  Finally, we select the adversarial example $x' \in X'$ with the highest ET as the final output ($x^*$). This sample has the highest estimated likelihood of successfully transferring to an unknown target model $f$.
\end{enumerate}

\begin{algorithm}[t]
\SetAlgoLined
\DontPrintSemicolon
\KwIn{Target image $x$, substitute model $f'$, set of substitute models $\mathcal{F}$, sampling function $S$}
\KwOut{Adversarial example $x^*$}

$X' \gets \emptyset$ \;
\For{$i \gets 1$ to $n$}{
    $x_i \gets S(x)$ // generate a perceptually equivalent sample \;
    $x'_i \gets \text{Attack}(f', x_i)$  // generate adversarial example \label{line:alg}\; 
    $X' \cup x'_i$
}

$x^* \gets \underset{x' \in X'}{\arg\max} ~\text{ET}_\mathcal{F}(x')$ // find the most transferable sample\;

\Return $x^*$ 
\caption{Perceptual Exploration Attack}\label{alg:peas}
\end{algorithm}

\subsection{Sampling Functions}
The strength of PEAS depends on how well the sampling function $S$ perturbs the robust features in $x$. In this work, we propose two basic sampling functions that can be used with PEAS: $S_1$ and $S_2$. Let $A$ be a set of augmentation algorithms, each configured to perform a subtle augmentation (e.g., one may rotate an image randomly on the range $[-2,2]$ degrees). The function $S_1$ applies a random augmentation from $A$ to $x$. $S_2$ applies all augmentations in $A$ for each  to $x$ (e.g., the `Mix' example from Fig. \ref{fig:augnorms}). Overall, the tradeoff between the two is that $S_1$ is stealthier while $S_2$ is more effective at exploring transferable versions of $x$.

\subsection{Improving Black Box Attacks with PEAS}\label{subsec:bbpeas}
PEAS can be seen as a method for moving samples closer to common model boundaries. Line \ref{line:alg} in Algorithm \ref{alg:peas} enables us to apply this strategy to existing black box attack algorithms; increasing their likelihood of success. In general, black box attacks either utilize substitute model(s) to create $x'$ (e.g., \cite{liu2016delving,dong2019evading}), query the victim $f$ to refine $x'$ (e.g., \cite{chen2017zoo,guo2019simple}) or do both (e.g., \cite{lord2022attacking,cai2022blackbox}). We'll discuss how PEAS can be integrated in all cases:

\begin{description}[itemsep=0pt]
    \item[Attacks which use Substitute Models:] To use PEAS in these attacks,  all we need to do is replace ``Attack'' on line \ref{line:alg} in Algorithm 1 with the chosen attack. By doing so, we are effectively using the other black box attack as a means for searching for samples with better transferability.

    \item[Attacks which Query the Victim:] In this setting, we do not want to execute the attack as part of PEAS since this would result in an increased query count on the victim (which is not covert) and would lead to an \textit{overt} adversarial example since we'd be repeatedly applying augmentations to the same sample. To resolve this, we \textit{first} execute PEAS and then pass $x^*$ to the other black box attack -giving it a better starting point. Doing so not only increases the likelihood of success but can also reduce the query count.
\end{description}


\section{Evaluation}\label{sec:eval}
In this section, we evaluate the performance of PEAS as a `plug-and-play' strategy for improving existing black box attacks. We also investigate why the attack is effective through an ablation study. To reproduce our work, the reader can find the source code to PEAS online.\footnote{https://github.com/BarAvraha/PEAS/tree/main}

\subsection{Experiment Setup}

\textbf{Attack Model.} We assumed the following attack model in our experiments: the adversary is operating in a black box setting where there is no knowledge of the target model's parameters or architecture. We assume that the adversary knows the training data's distribution, as commonly assumed in other works \cite{tramer2016stealing,zhu2021hermes, qin2023training, cai2022blackbox}. In our setting, the attacker wishes to perform an untargeted attack where the objective is to cause an arbitrary classification failure: $f(x')\neq y$. Although PEAS can be easily adapted to the targeted setting, we leave this analysis to future work.

\textbf{Datasets.} To evaluate PEAS, we used two well-known benchmark datasets: \texttt{CIFAR-10} and \texttt{ImageNet}.  \texttt{CIFAR-10} is an image classification dataset consisting of 60K images with 10 classes having a resolution of 32x32.
\texttt{ImageNet} contains approximately 1.2M images with 1000 classes rescaled to a resolution of 224x224. For both datasets, we used the original data splits. As mentioned, we used the same training data for $f$ and $f'$  following the work of other black box attack papers \cite{tramer2016stealing,zhu2021hermes, qin2023training, cai2022blackbox}. Following the setup of other similar works (e.g., . \cite{cai2022blackbox, ge2023boosting, long2022frequency}) we evaluated 1000 random samples from the testing data of each dataset. To avoid bias, we only used samples that were correctly classified by $f$.

\textbf{Architectures.}
In our experiments, we used ten different architectures, five for each dataset. These architectures were used to demonstrate that PEAS works in a black box setting (no knowledge of the architecture of $f$) and under different configurations. We used pretrained models for both datasets.\footnote{CIFAR-10 models: \url{https://github.com/chenyaofo/pytorch-cifar-models}\\ImageNet models: \url{https://pytorch.org/vision/stable/models.html}} 
The architectures used for \texttt{CIFAR-10} were: \texttt{Resnet-20}, \texttt{VGG-11}, \texttt{RepVGG-a0}, \texttt{ShuffleNet v2-x1-5} and \texttt{Mobilenet v2-x0-5}.
The architectures used for \texttt{ImageNet} were: \texttt{DenseNet-121}, \texttt{Efficientnet}, \texttt{Resnet18}, a vision transformer (\texttt{ViT}) and a Swin transformer (\texttt{Swin-s}). 

These architectures were selected to capture diversity in deep learning models. For example, \texttt{ViT} applies the principles of transformer models, primarily those used in natural language processing, to image classification tasks. It treats image patches as sequences, allowing for global receptive fields from the outset of the model.

\textbf{PEAS Setup.}
In our experiments, we used both the $S_1$ and $S_2$ sampling functions. $S_2$ is used when the sampling function is not indicated. In all cases, we set the number of augmentations per input image ($n$) to 200. For the set of augmentations $A$, we used the following transformations:
\begin{itemize}
    \item \textbf{RandomAffine} Random rotations (between -2 to 2 degrees for \texttt{ImageNet}, -4 to 4 for \texttt{CIFAR-10}) and translations up to 10\% of image dimensions.
    \item \textbf{ColorJitter} Random adjustments with increments of 0.05 for brightness, contrast, saturation, and hue.
    \item \textbf{RandomCrop} Random crops to 224x224 pixels with 10 pixels padding for \texttt{ImageNet} and 32x32 with 3 pixels padding for \texttt{CIFAR-10}.
    \item \textbf{GaussianBlur} Blur with a kernel size of 3 and 1.9 for \texttt{ImageNet} and \texttt{CIFAR-10} respectively.
    \item \textbf{RandomAdjustSharpness} A sharpness factor of 2 and 1.5 for \texttt{ImageNet} and \texttt{CIFAR-10} respectively -applied universally.
    \item \textbf{RandomAutocontrast} Auto contrast applied at random 50\% of the time.
\end{itemize}

To set up an attack with the five architectures (per dataset), we used one as $f$, one as $f'$, and the remaining three as $\mathcal{F}$. In all of our experiments we evaluate every possible combination.
Because of the black box assumption, $f' \neq f$ in every setting. 

\textbf{Baselines \& Metrics.}
As a baseline for performance, we compare PEAS two different transfer-based black box attack strategies: attacking without ranking (naive transferability from $f'$ to $f$) and attacking with ranking (the Vanilla ET ranking technique from equation (\ref{eq:het}) \cite{levy2022transferability}). The Vanilla ranking approach is equivalent to using PEAS with a sampling function that simply adds noise to $x$ from within an $\epsilon$-ball. 

We also evaluate how much PEAS boosts the performance of five existing black-box attacks: Basic Transfer Attack (BTA), which uses PGD on a surrogate to create \( x' \); FGSM-TIMI \cite{dong2019evading}, similar to BTA but with input diversity; SimBA \cite{guo2019simple}, which queries the victim for feedback; and two recent attacks, PGN \cite{ge2023boosting} and SSA \cite{long2022frequency} which enhance sample transferability by averaging gradients from multiple samples and applying spectrum transformations respectively.
We denote a boosted attack as X-PEAS where X is the name of the attack algorithm which we are boosting (e.g. BTA-PEAS).

We set $\epsilon$ to $2/255=0.0078$ for \texttt{CIFAR-10} and to $12.75/255=0.05$ for \texttt{ImageNet} based on other black box attack papers (e.g., \cite{rnd-blackbox-defense-nips2021,cgattack-cvpr2022}). 
Finally, we performed an ablation study and hyperparameter evaluation to analyze how each component of PEAS contributes to the attack's performance.
To measure performance, we calculate the attack success rate (ASR), which is the ratio of samples that are misclassified by the victim model.  


\subsection{Baseline Evaluation}
\textbf{Boosting with Ranking.} In Table \ref{tab:baseline}, we compare (1) the performance of ranking random starts using noise \cite{levy2022transferability} (Vanilla) and ranking random augmentations (BTA-PEAS). In both cases we use PGD to generate the perturbations on the starting points. The lower bound is BTA (basic transferability attack) and the upper bound is the simulated case where a `perfect ranking algorithm' is used in PEAS with $S_2$. 

\begin{figure*}
    \includegraphics[width=\textwidth]{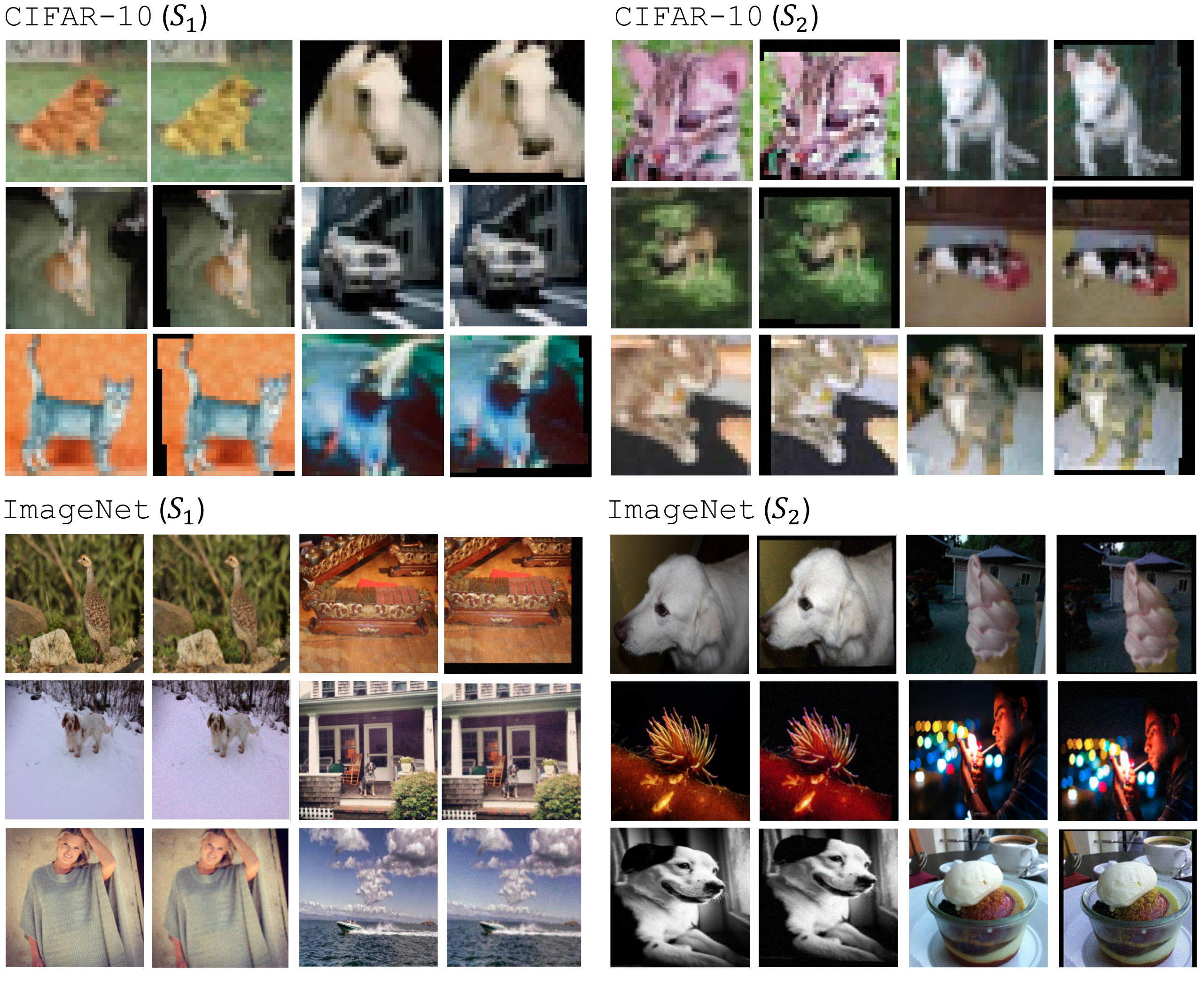}
    \caption{Sample images before ($x$) and after ($x^*$) the application of the BTA-PEAS attack using two different sampling functions, $S_1$ and $S_2$. The left image is $x$ (correctly classified by $f$) and the right image is the black box adversarial example (misclassified by $f$).}\label{fig:samples}
\end{figure*}

The table shows that Vanilla ranking (ranking random starts) is ineffective, as seen by its comparison to the lower bound (BTA). In contrast, BTA-PEAS is much more effective, achieving an average improvement in attack success rates of 1.7x with $S_1$ and 2.5x with $S_2$, with some cases reaching a 6.3x gain. This validates our hypothesis that Vanilla's additive noise does not effectively perturb transferability-critical features, while PEAS targets them effectively with augmentations (see Fig. \ref{fig:samples} for examples). Although PEAS performs significantly better, there is still room for improvement, as indicated by the upper bound. Enhancing the ET ranking algorithm and developing better sampling functions can achieve this.

In summary, PEAS's augmentation ranking strategy significantly outperforms both baseline transferability and Vanilla ranking, highlighting the importance of targeting robust features for improved adversarial transferability.

\begin{table}[t]
    \caption{The attack success rate of BTA-PEAS compared to the Vanilla ranking approach, for different combinations of architectures for the victim and substitute models. The lower bound (left) is basic transferability from $f'$ to $f$, and the upper bound simulates the result of a perfect ranking algorithm.}
    \label{tab:baseline}
    \centering
    \includegraphics[width=0.8\columnwidth]{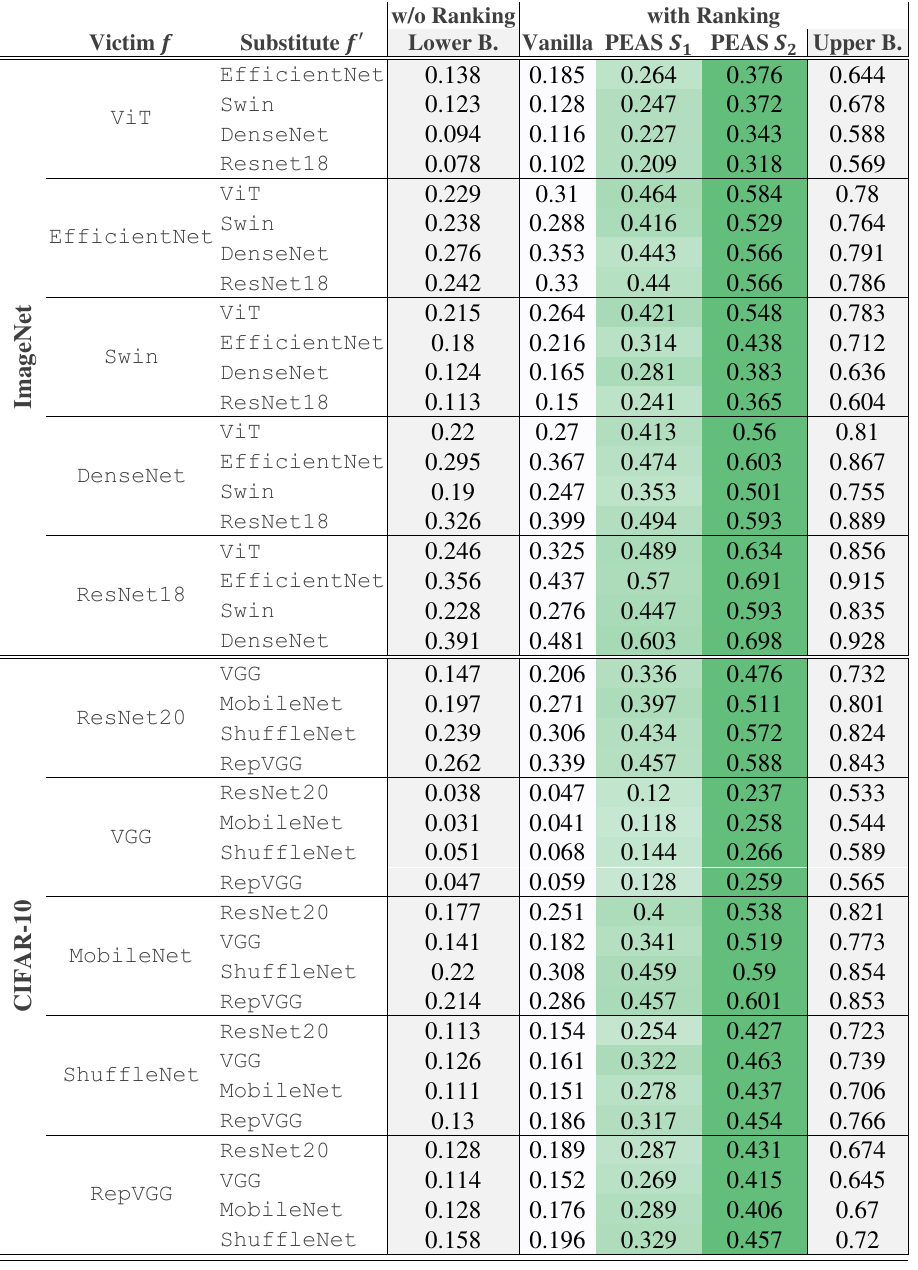}
\end{table}

\textbf{Boosting Black Box Attacks.} In Table \ref{tab:timi_simba}, we compare the performance of different attack strategies before and after applying PEAS. The strategies are (1) a basic transfer attack using a surrogate (BTA), (2) a transfer attack using input diversity (FGSM-TIMI), and (3) an iterative attack using query feedback (Simba). Here, sampling strategy $S_2$ is used.
The results show that by boosting the basic transfer attack, PEAS can obtain a performance 7.4x and 1.6x better than TIMI and SimBA respectively, on average. 
We can also see that even when input diversity (TIMI) is used or when the attack is querying the black box victim (Simba), PEAS can increase the ASR by a factor of 1.35. 


In Table \ref{tab:pgn_ssa}, we show the performance of two state-of-the-art black box attacks (PGN and SSA) and show how PEAS boosts their ASR for different epsilon budgets. Note that an epsilon of 25.5/255 is not considered stealthy. We also present the performance of the simple BTA attack as reference. Both PGN-PEAS and SSA-PEAS outperform their original versions by a significant margin.

Overall, these results demonstrate that PEAS can be effectively leveraged as a performance-enhancing strategy for various existing black box attacks, including modern ones.

\begin{table}[t]
    \centering
        \caption{The performance of three different black box attack strategies with and without boosting from PEAS. All results are presented in ASR averaged across all substitute models.}
    \label{tab:timi_simba}
    \includegraphics[width=0.7\columnwidth]{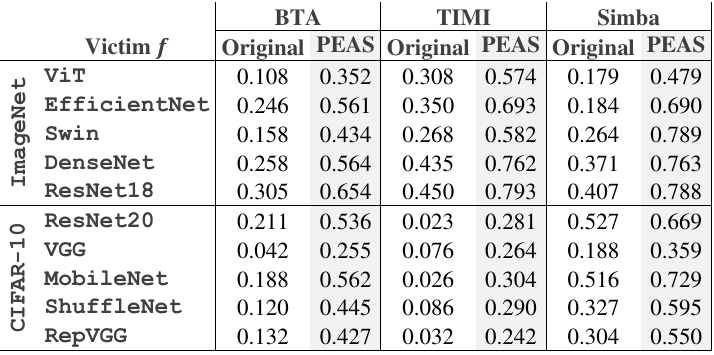}
\end{table}

\begin{table}[t]
    \centering
        \caption{The performance of two state of the art black box attacks (PGN and SSA) with and without boosting from PEAS for different epsilon. Note: an epsilon of 25.5/255 is not considered stealthy.}
    \label{tab:pgn_ssa}
    \includegraphics[width=0.8\columnwidth]{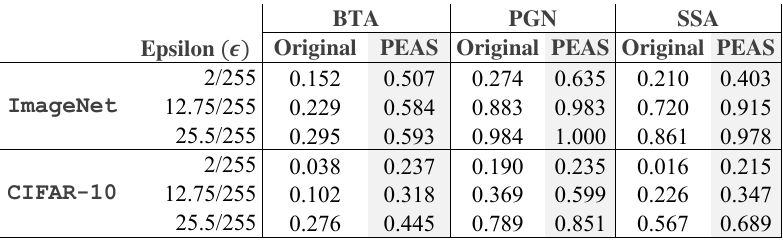}
\end{table}

 \begin{table*}[t]
     \centering
          \caption{An ablation study of the PEAS algorithm. Here, the basic transfer attack (BTA) is being boosted. Each column represents a different strategy for selecting the attack sample from the samples generated by $S$. `Filtered' means that we do not include augmentations of $x$ that fool $f$. All results are presented as ASR averaged across all substitute models. Shading indicates the best results per row.}
     \label{tab:ablation}
     \includegraphics[width=\textwidth]{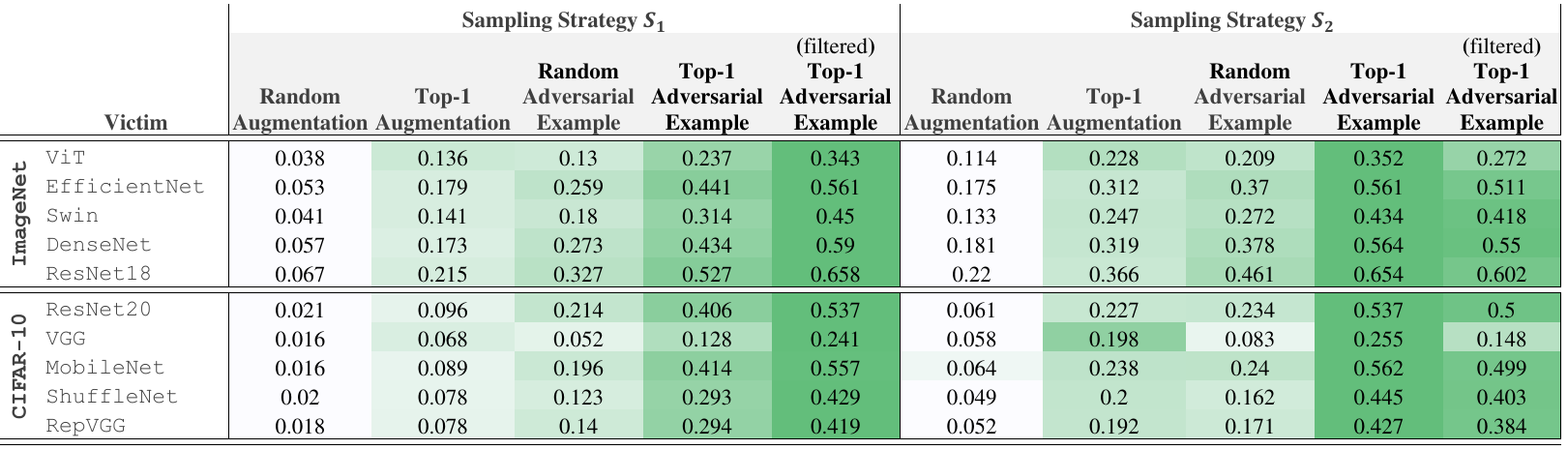}
 \end{table*}

\subsection{Ablation Study}
\textbf{Algorithm Components.} 
In Table \ref{tab:ablation}, we investigate the
contribution of each component of PEAS in selecting the best sample (augmentation) from $S$. For example, `Top-1 Adversarial Example' is the proposed PEAS algorithm where we first \textit{attack} each sample in $S$ and then \textit{select the top sample} using ET with $\mathcal{F}$. Here we are boosting the basic transfer attack (BTA).

Our first observation is that PEAS succeeds not because augmentations cause misclassifications, but because they provide better starting points for attacks. This can be seen by contrasting the column ``Random Augmentation'' (simply using augmentations as the attack) to ``(filtered) Top-1 Adversarial Example'' (where we only use samples that don't cause a natural misclassifcation). Regardless, in a real attack some augmentations may increase the ASR due to natural misclassifications. However, we argue that these are legitimate perturbations an adversary can use, as they are subtle. The key contribution is selecting the best one to use. 

Our second observation highlights the role of augmentation in transferability: randomly selected augmented samples yield poor performance, but the top-1 augmented sample performs decently. This demonstrates that (1) ET works on augmented samples, and (2) PEAS attacks can be performed without adversarial perturbations. However, adding a perturbation on top of the augmented sample creates a more effective attack, as augmentation positions the sample advantageously, and the perturbation pushes it over the boundary. Therefore, both augmentations and perturbations are necessary for a strong attack in PEAS.

We also compare the performance of single augmentations applied randomly ($S_1$) versus a mix of augmentations ($S_2$). Results in Table \ref{tab:ablation} show $S_2$ is inherently more robust. For $S_1$, the average ASR increases when deceptive augmented samples are removed, suggesting these augmentations often retreat across the decision boundary due to misaligned gradients. In contrast, the ASR for $S_2$ decreases after similar filtering, indicating $S_2$ augmentations provide more reliable starting points that extend deeper beyond the decision boundary, resulting in more stable attacks. Thus, a mix of augmentations, as in $S_2$, is preferred for its effectiveness and robustness.

\textbf{Hyperparameters.}
One of the key hyperparameters of PEAS is $n$, the exploration size, which determines how many versions of $x$ are produced using the sampling function $S$ before ranking.

\begin{figure}[t]
     \centering
     \includegraphics[width=0.8\columnwidth]{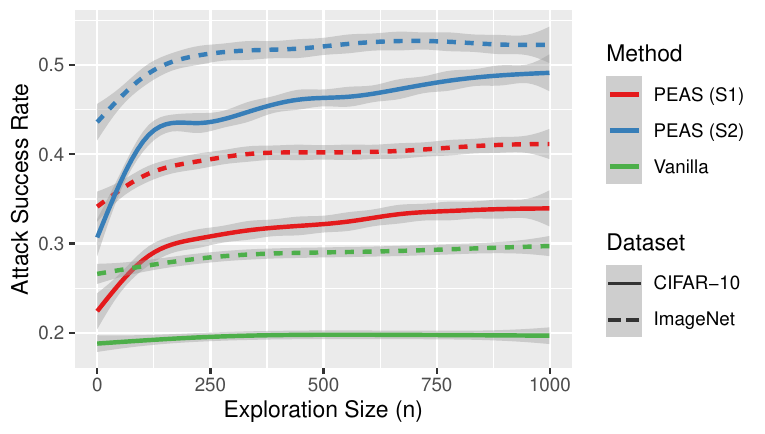}
     \caption{The effect of the exploration size $n$ on the performance of BTA-PEAS and the Vanilla ranking strategy. The grey margin captures the confidence interval for $p=0.99$.}
     \label{fig:counts}
     \vspace{-1em}
 \end{figure}

Figure \ref{fig:counts} compares the performance of BTA-PEAS to Vanilla Ranking for increasing values of $n$. The plot shows that even with $n=1$, BTA-PEAS outperforms Vanilla Ranking in generating adversarial examples, supporting our finding that transferable adversarial examples can be crafted through subtle augmentations alone. This presents a challenge for defenses designed to detect or mitigate adversarial noise \cite{serban2020adversarial}.

BTA-PEAS maintains a significant performance advantage over Vanilla Ranking across all tested values of $n$. The performance of PEAS converges around an exploration size of $n=200$, indicating that only 200 samples are needed to sample our distribution of augmentations. For an analysis of the effect of $n$ on each augmentation in $A$, please see the supplementary material.


\begin{figure}[ht]
    \centering
    \includegraphics[width=0.8\columnwidth]{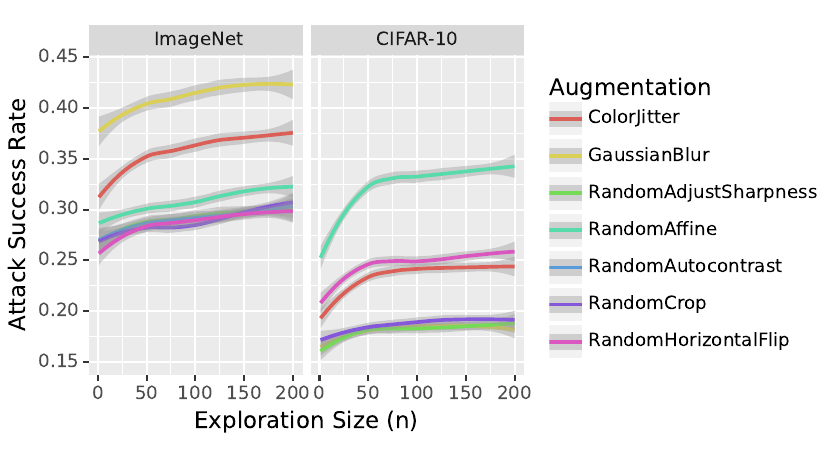}
    \caption{The performance of each augmentation in BTA-PEAS as a function of the number of instances of \(x\) created during the exploration step. The gray margin captures the confidence interval for $p=0.95$.}
    \label{fig:augmentations}
\end{figure}

\textbf{Effectiveness of Augmentations.}
Figure \ref{fig:augmentations} evaluates the effectiveness of various augmentations in PEAS. The results show that Gaussian Blur and Random Affine transformations are most effective for high and low-resolution datasets, respectively. Gaussian Blur is effective on high-resolution images by removing fine details, thus forcing the attack to focus on robust features. Conversely, Random Affine transformations significantly impact low-resolution images by altering the alignment and appearance of robust features, creating a greater challenge for spatial generalization of the model $f$.

\begin{figure}[ht]
    \centering
    \includegraphics[width=0.8\columnwidth]{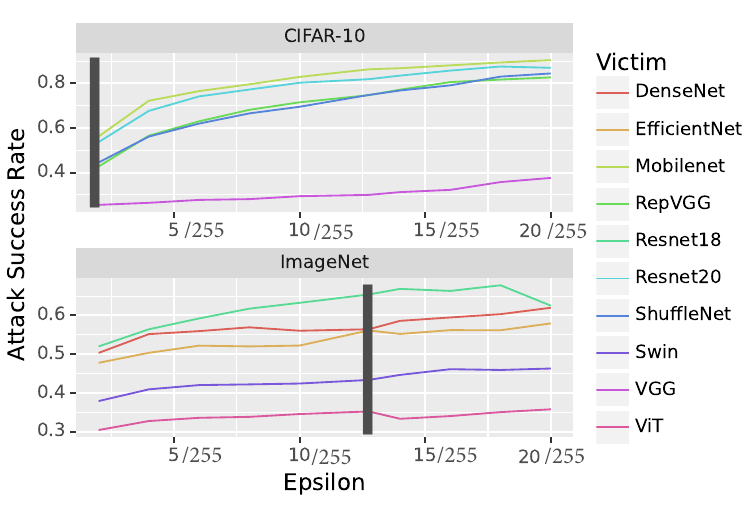}
    \caption{Effect of increasing the $\epsilon$-budget in BTA-PEAS for each victim architecture $f$. Results are averaged across different substitute models where $f \neq f'$.}
    \label{fig:epsilon}
\end{figure}

\textbf{Impact of $\epsilon$-Budget.} 
Figure \ref{fig:epsilon} illustrates the relationship between the $\epsilon$-budget (perturbation size) and the attack success rate of BTA-PEAS across different victim architectures $f$. The vertical bars on the figure indicate the standard $\epsilon$ values used for all attacks in the main paper, providing a reference point for the typical perturbation strengths considered in our experiments.

As expected, the attack success rate consistently increases with a larger $\epsilon$, as this allows the adversarial perturbation to become more pronounced. A higher $\epsilon$ gives the adversary more room to introduce changes, helping the perturbation traverse decision boundaries that may be misaligned between the substitute model $f'$ and the target model $f$, supporting the observations of \cite{demontis2019adversarial}. 

Interestingly, while this trend is observed across all victim architectures, the rate of increase varies depending on the architecture’s robustness to perturbations. For example, architectures like Vision Transformers (ViT) exhibit a more gradual improvement compared to convolutional networks like ResNet, which see more immediate gains as $\epsilon$ grows. This suggests that different model architectures might have distinct sensitivities to perturbation sizes, and BTA-PEAS is particularly effective at exploiting those that rely more heavily on non-robust features for classification.

\textbf{Complexity.}
While PEAS requires the execution of an attack algorithm $n$ times per sample, we argue that this is an acceptable cost, depending on the scenario. Consider where the adversary must succeed on the first try (e.g., evading surveillance, bank fraud, tampering with medical scans) or perform minimal attempts (queries) to avoid detection. In this cases, spending even a day to make one sample is a reasonable price to avoid being caught. With BTA-PEAS, we found that on an ADA RTX6000 GPU it takes 3 minutes to make an adversarial example for CIFAR-10 and 5 minutes for ImageNet (with a batch size of one).

\section{Conclusion}
In conclusion, our Perception Exploration Attack Strategy (PEAS) can boost black box adversarial attacks by finding an ideal perceptually equivalent starting point which enhances transferability. 
This work both introduces an effective attack strategy and deepens our understanding of adversarial transferability, highlighting perceptual equivalence as a powerful tool in adversarial machine learning.


\bibliographystyle{ACM-Reference-Format}
\bibliography{egbib}

\end{document}


\title{Supplementary Material for \\ PEAS: A Strategy for Crafting Transferable Adversarial Examples}



\renewcommand{\shortauthors}{XXXX et al.}

This document provides supplementary material for our paper. It includes code access, additional experimental results, and visualizations to complement the main manuscript.

\section{Reproducibility}
We are dedicated to the principles of transparency and reproducibility in research. To support this, we have made our codebase available, along with instructions for running the main evaluation experiment discussed in the paper. Please feel free to use our code to verify and experiment with PEAS.

The code can be found in a colab notebook here:\\
\url{https://colab.research.google.com/drive/1Y6nIXxgG2BhwE4SNUCiObcSEboi1QY3Z?usp=sharing}\\

The example in the notebook uses PEAS to boost the BTA attack on a CIFAR-10 model. The models/dataset can be changed accordingly. The code can be used to boost any non-query based attack (e.g., BTA, PGN, TIMI, ...) in a plug-and-play manner. Please note that results will vary slightly between runs due to the random processes (augmentations and selection of images).  

\section{Exploration Size Analysis}
\begin{figure*}[ht]
    \centering
    \includegraphics[width=\textwidth]{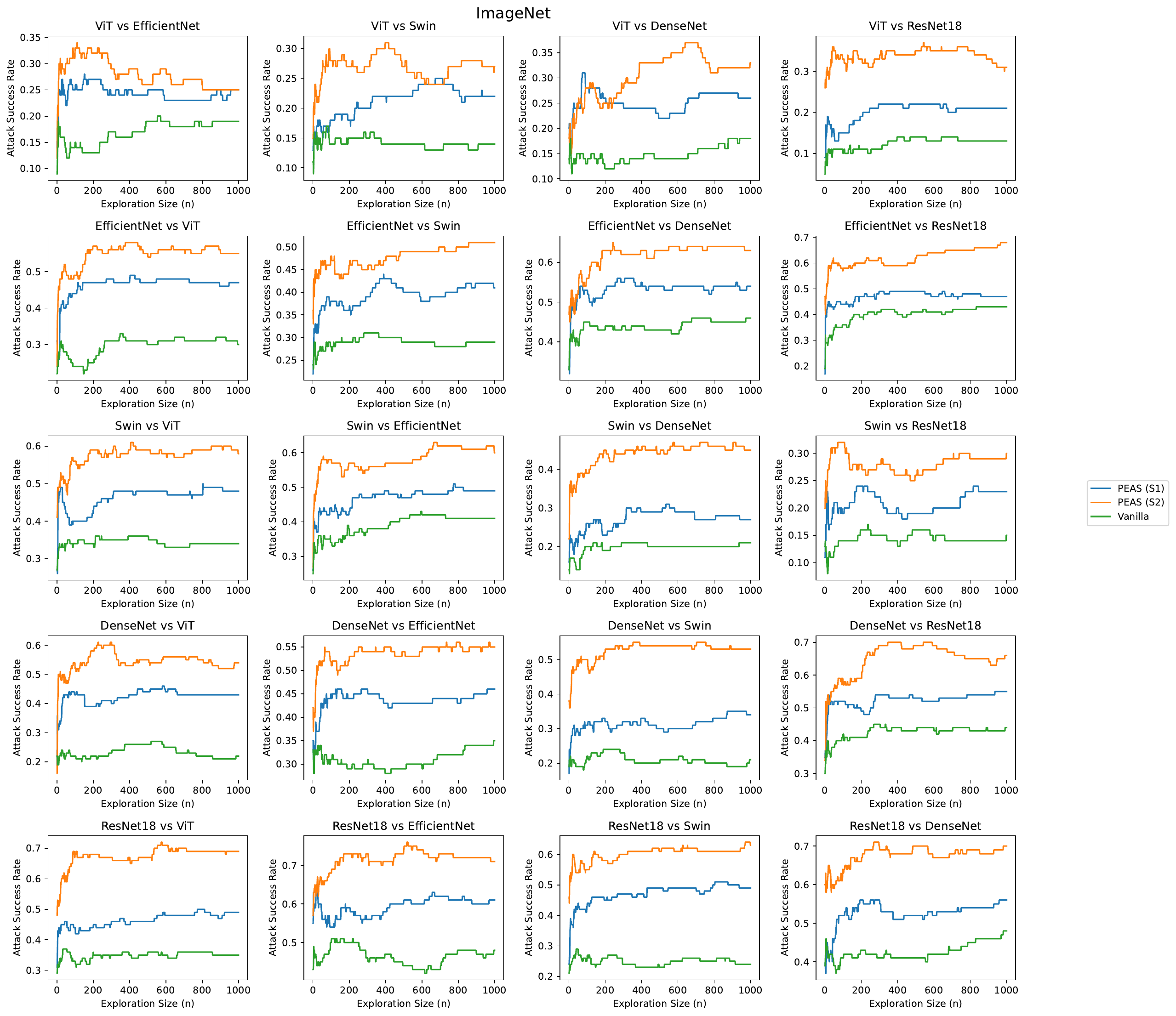}
    \caption{A detailed exploration size analysis of the parameter $n$ for the ImageNet dataset. Here we compare the performance of PEAS to the Vanilla ranking strategy across all combinations of victim and surrogate models.}
    \label{fig:sample_size_imagenet}
\end{figure*}

\begin{figure*}[ht]
    \centering
    \includegraphics[width=\textwidth]{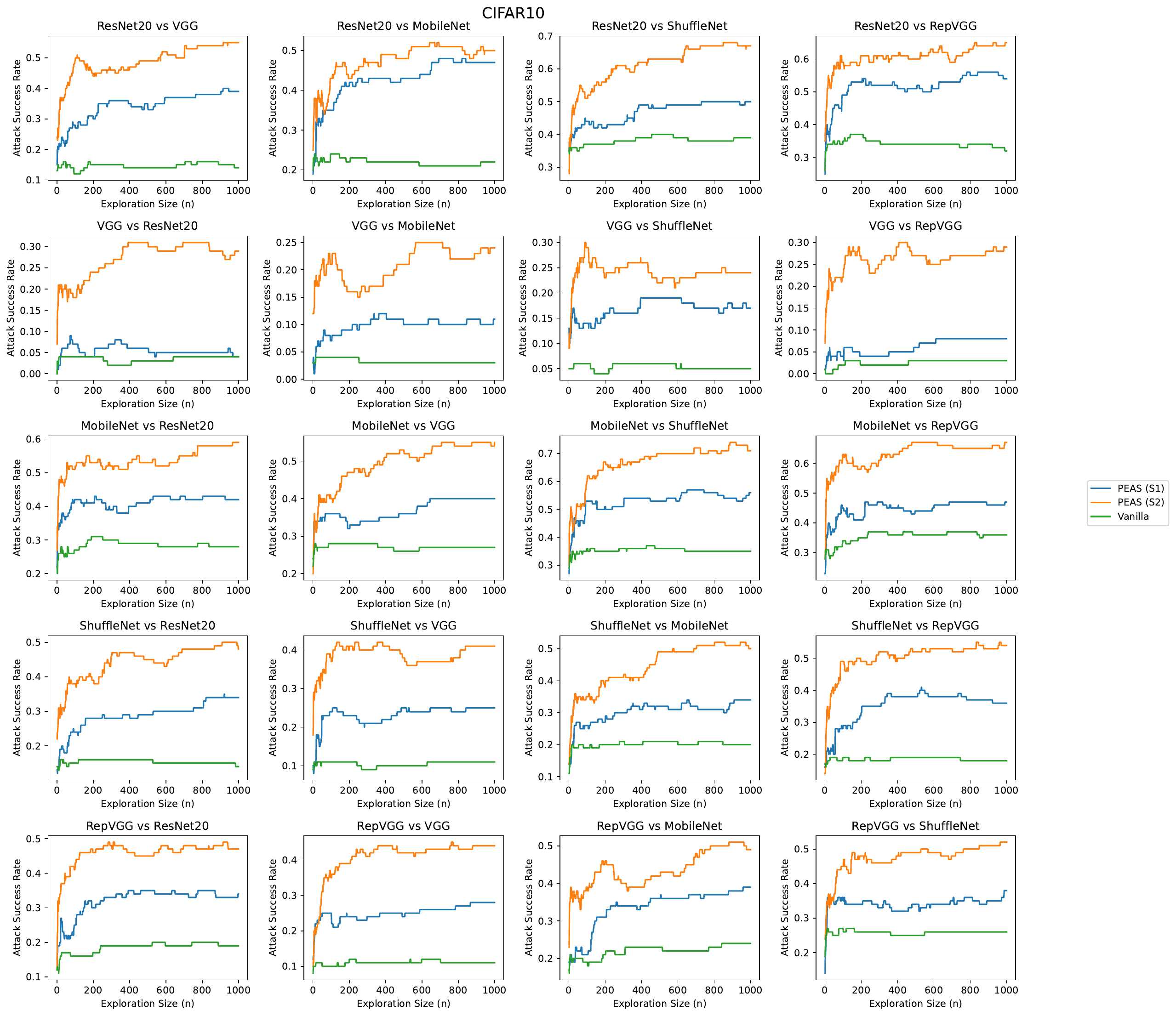}
    \caption{A detailed exploration size analysis of the parameter $n$ for the CIFAR-10 dataset. Here we compare the performance of PEAS to the Vanilla ranking strategy across all combinations of victim and surrogate models.}
    \label{fig:sample_size_cifar}
\end{figure*}

We provide a more detailed view of how the exploration size $n$ affects the performance of BTA-PEAS. In Fig. 4 of the main paper, we averaged across every surrogate-victim combination. Here, we provide the results for all combinations of victim and surrogate models and compare BTA-PEAS to the Vanilla ranking strategy. Figure \ref{fig:sample_size_imagenet} is for ImageNet and Figure \ref{fig:sample_size_cifar} is for CIFAR-10.

